\documentclass[10pt,twocolumn,letterpaper]{article}

\usepackage{iccv}
\usepackage{times}
\usepackage{epsfig}
\usepackage{graphicx}
\usepackage{amsmath}
\usepackage{amssymb}

\usepackage{array}
\usepackage{xspace}
\usepackage{enumitem}
\usepackage{subfigure}
\usepackage{multirow}
\usepackage{stfloats}
\usepackage{caption}
\usepackage{booktabs}

\usepackage[breaklinks=true,bookmarks=false]{hyperref}

\iccvfinalcopy 

\def\iccvPaperID{56} 

\ificcvfinal\fi

\begin{document}

\newcommand{\proposal}{{\textit{SlAction}}\xspace}
\newcommand{\hs}[1]{{\color{black}#1}}

\title{SlAction: Non-intrusive,  Lightweight Obstructive Sleep Apnea Detection using Infrared Video}

\author{You Rim Choi$\dagger$$^1$ \hspace{.3cm} Gyeongseon Eo$\dagger$$^1$ \hspace{.3cm} Wonhyuck Youn$^3$ \\
Hyojin Lee$^1$ \hspace{.3cm} Haemin Jang$^1$ \hspace{.3cm} Dongyoon Kim$^4$ \hspace{.3cm} Hyunwoo Shin*$^{2,3}$ \hspace{.3cm} Hyung-Sin Kim*$^1$\\
Seoul National University$^1$ \hspace{.3cm} Seoul National University College of Medicine$^2$\\
OuarLab$^3$ \hspace{.3cm} Columbia University$^4$\\
{\tt\small \{yrchoi, eks104, sarahhyojin, haemin.jang, charlie, hyungkim\}@snu.ac.kr}\\
{\tt\small wonhyuck.yoon@ouarlab.com \hspace{.3cm} dk3315@columbia.edu}
}

\maketitle
\ificcvfinal\thispagestyle{empty}\fi

\def\thefootnote{$\dagger$}\footnotetext{These authors contributed equally to this work}\def\thefootnote{\arabic{footnote}}
\def\thefootnote{*}\footnotetext{Corresponding authors}\def\thefootnote{\arabic{footnote}}

\begin{abstract}
Obstructive sleep apnea (OSA) is a prevalent sleep disorder affecting approximately one billion people worldwide. 
The current gold standard for diagnosing OSA, Polysomnography (PSG), involves an overnight hospital stay with multiple attached sensors, leading to potential inaccuracies due to the first-night effect.
%
To address this, we present \proposal, a non-intrusive OSA detection system for daily sleep environments using infrared videos.
Recognizing that sleep videos exhibit minimal motion, this work investigates the fundamental question: ``\textit{Are respiratory events adequately reflected in human motions during sleep?}''
%
Analyzing the largest sleep video dataset of \hs{5,098 hours}, we establish correlations between OSA events and human motions during sleep.
%
Our approach uses a low frame rate (2.5 FPS), a large size (60 seconds) and step (30 seconds) for sliding window analysis to capture slow and long-term motions related to OSA. Furthermore, we utilize a lightweight deep neural network for resource-constrained devices, ensuring all video streams are processed locally without compromising privacy.
Evaluations show that \proposal achieves an average F1 score of \hs{87.6\%} in detecting OSA across various environments. 
%
Implementing \proposal on NVIDIA Jetson Nano enables real-time inference (\hs{$\sim$3 seconds} for a 60-second video clip), highlighting its potential for early detection and personalized treatment of OSA.


\end{abstract}

\vspace{-2ex}
\section{Introduction}

Obstructive sleep apnea (OSA) is one of the most common sleep disorders characterized by recurrent partial (hypopnea) or complete (apnea) obstructions of the upper airway, resulting in temporary cessation of breathing~\cite{strollo1996obstructive, gottlieb2020diagnosis}. 
OSA affects approximately one billion people, comprising about 14\% of the global population aged 30 to 69 years~\cite{benjafield2019estimation}. Despite its high prevalence~\cite{peppard2013increased, punjabi2008epidemiology, mirrakhimov2013prevalence, heinzer2015prevalence, franklin2015obstructive, senaratna2017prevalence}, 
\textbf{early diagnosis of OSA is challenging} due to a lack of awareness among individuals about its symptoms such as snoring and breathing cessation during sleep, leading to delayed medical consultation~\cite{mcnicholas2008diagnosis, reuveni2004awareness, mcnicholas2018challenges, randerath2018challenges}.

If an individual recognizes their OSA symptoms and seeks professional sleep medicine, Polysomnography (PSG) is considered the gold standard for sleep evaluation. PSG involves the patient spending a night at a specialized sleep laboratory with a dozen sensors attached to record various signals. Physicians then manually analyze the several hours of recorded signals
~\cite{RUNDO2019381}. The primary clinical metric for characterizing OSA, apnea-hypopnea index (AHI), is derived from PSG measurements~\cite{american1999sleep, adult2009clinical}. 
However, PSG-based OSA diagnoses can be inaccurate due to  two major factors: 
(1) Attaching sensors and sleeping in an unfamiliar environment can lead to the \textbf{first-night effect}, causing significant differences between measurements from first night and those during daily sleep.
(2) The variability of respiratory events causes substantial fluctuations in OSA severity from night to night, but PSG is typically measured for only \textbf{single night} due to its complex and costly setup.

Several studies have explored daily remote sleep monitoring systems to facilitate early detection of OSA and comprehensive diagnostic evaluations at healthcare facilities~\cite{van2011objective, park2019smart}. 
These systems have utilized various IoT sensors 
~\cite{kim2020iot, castillo2021detection, fino2019monitoring, ko2015consumer, davidovich2016sleep, hafezi2020sleep, lin2017iot, nandakumar2015contactless}, and wearable devices~\cite{mantua2016reliability, chen2021apneadetector, kim2022diagnostic}. 
However, they have shown limited correlations with PSG and lower levels of accuracy, often evaluated with small sample sizes of fewer than 40 subjects, restricting the generalizability of the results.

This work explores the potential of \textit{infrared sleep videos} for non-intrusive OSA monitoring during daily sleep. 
The application scenario is as follows: A video camera is installed on the wall or ceiling, positioned away from the bed to ensure \textit{uninterrupted sleep} for the subject. The video streams are processed locally in real-time to calculate the AHI automatically, ensuring privacy. The device can aggregate AHI results over multiple nights, enabling self-diagnosis and issuing local alerts for medical attention. Additionally, AHI results can be sent to the doctor for remote diagnosis and personalized treatment.

\vspace{1ex}\noindent
\textbf{Challenges.} Characterizing OSA in infrared sleep videos presents unique challenges compared to other video-based action recognition tasks. 
(1) Sleep videos capture \textit{the most inactive moment} of human beings. While most video analytics aim to mimic natural human perception, observing these nearly static videos does not provide significant insight to humans. Therefore, sleep video analytics must \textit{surpass human perception}.
(2) It has not been scientifically studied if apnea and hypopnea events are reflected in subtle human motions during sleep, distinct enough for a deep neural network (DNN) to differentiate from other motions. 
(3) Infrared videos captured from a distance suffer from poor quality, and noise due to anonymization and bedding obscures the chest and abdominal area where respiratory movements occur. 
(4) While enabling local inference on resource-constrained devices is desirable for privacy preservation, DNNs for video understanding typically have large model sizes and substantial computational requirements.

\section{Preliminary Study}

\subsection{Dataset}

\begin{figure}[t]
  \centering
    \includegraphics[width=.95\linewidth]{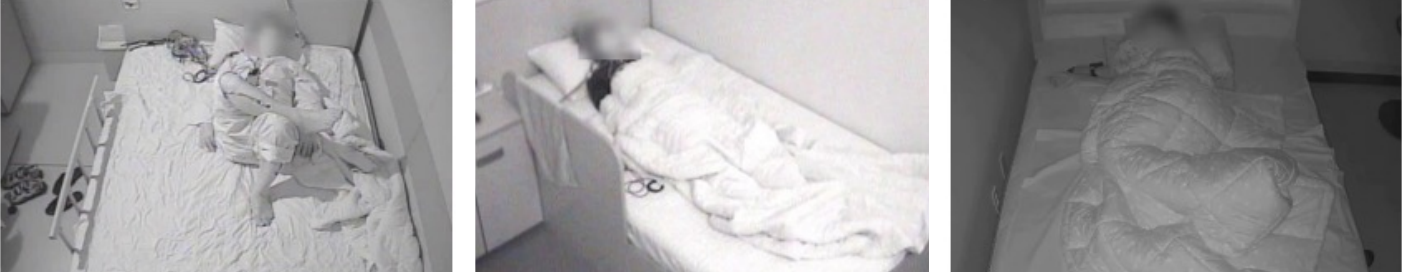}
    \vspace{-2ex}
  \caption{Example images in the dataset, showing various camera angles and room environments across hospitals, as well as diverse body sizes and sleep habits of patients.}
  \vspace{-2.5ex}
  \label{fig:sleep_video}
\end{figure}

We utilize the Infrared Sleep Video for Diagnosing Sleep Disorders dataset, which is the largest to our knowledge~\cite{AIHub_2022:online}. The dataset was collected from three hospitals (A to C), each with slightly different setups in terms of the equipment and examination environment, as depicted in Figure~\ref{fig:sleep_video} and Table~\ref{tab:collection_set}. 
we use infrared sleep videos from 729 patients, covering a diverse range of genders and ages spanning from teenagers to individuals over 80 years old.
Each video is an MP4 file with a resolution of 640$\times$480 and a frame rate of 5 FPS, with a duration of approximately 6 hours. 
The videos are synchronized with PSG data extensively annotated by sleep experts. 
To ensure privacy protection, face regions were anonymized by applying rectangle-based mosaic processing to the facial areas using object detection techniques.

\begin{table}[t]
    \centering
    \caption{Recording settings for video data}
    \label{tab:collection_set}
    \vspace{-2ex}
    \resizebox{\linewidth}{!}{ 
    {\tiny
    \begin{tabular}{lccc}
    \hline
        \toprule
                           & \textbf{Hospital A} & \textbf{Hospital B}  & \textbf{Hospital C} \\\hline
        Number of patients & 499         & 115       & 115 \\
        Shooting angle     & 30 degrees  & 45 degrees & 45 degrees \\
        Distance           & 3 m         & 3.5 m      & 3 m \\
        \bottomrule
        \hline
    \end{tabular}}
    }
    \vspace{-2.5ex}
\end{table}

\subsection{Key Insight}
\label{key_insight}

\begin{figure}[t]
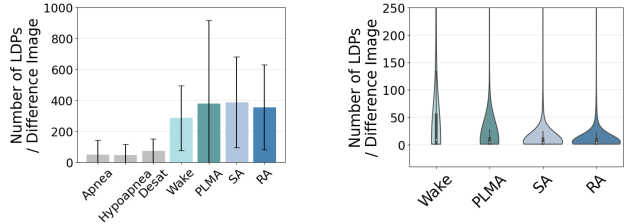

\begin{minipage}[t]{0.45\columnwidth}
    \subfigure[Average number of LDPs per difference image for entire sleep events]{
        \includegraphics[width=\textwidth]{figures/bar.pdf}
   \label{fig:Bar_all}
    }
\end{minipage}\hfill
\begin{minipage}[t]{0.45\columnwidth}
    \subfigure[Average number of LDPs per difference image during 10 seconds preceding each arousal or Wake]{
        \includegraphics[width=\textwidth]{figures/violin.pdf}
    \label{fig:PLM_RERA}
    }
\end{minipage}
\vspace{-3ex}
\caption{Comparison between various sleep events based on the number of LDPs per difference image}
\vspace{-3ex}
\label{fig:LDP}
\end{figure}

\begin{figure}[t]
  \centering
    \includegraphics[width=.75\linewidth]{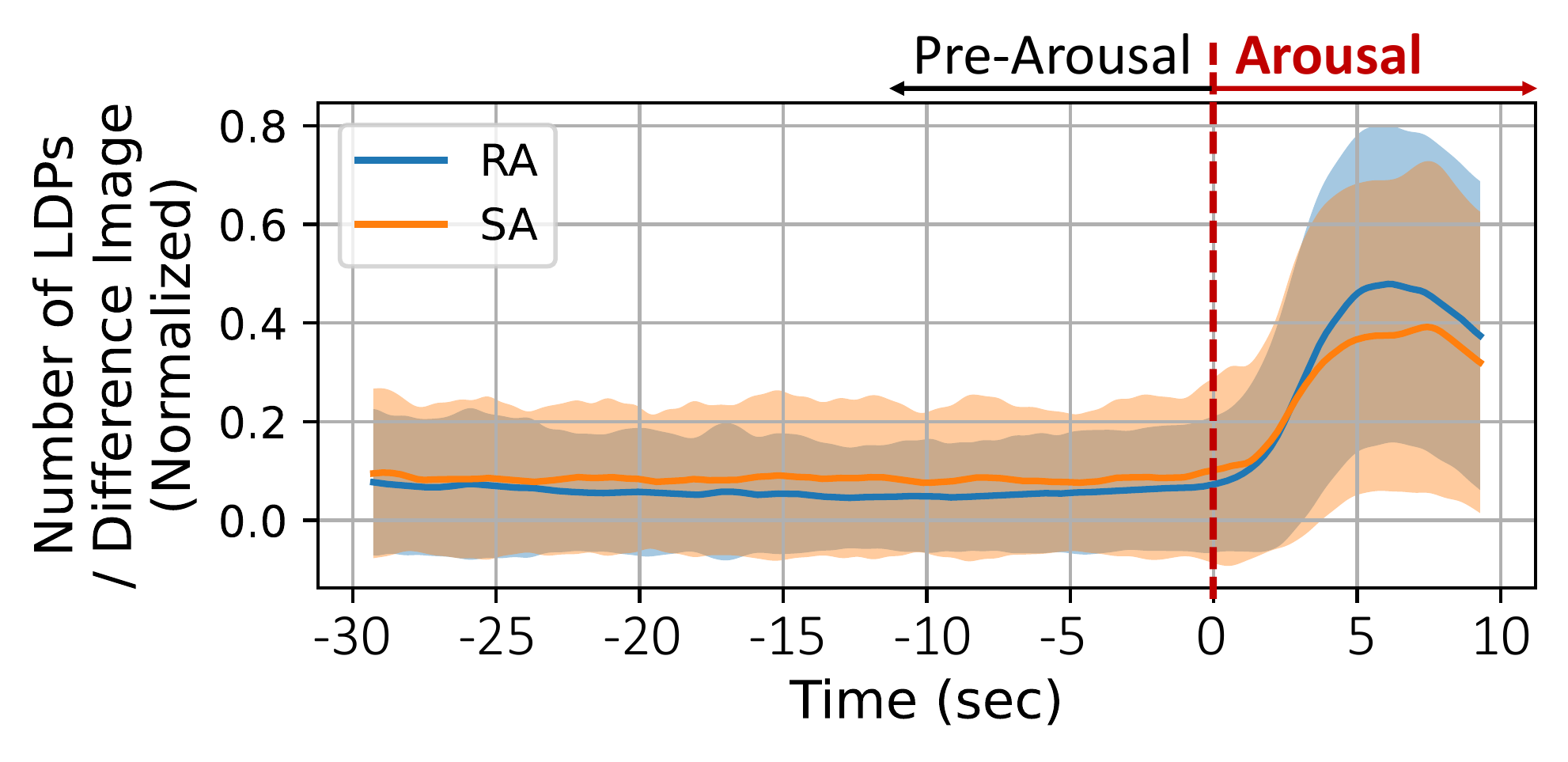}
    \vspace{-3ex}
  \caption{Evolution of LDP counts per difference image for respiratory arousal (RA) and spontaneous arousal (SA).}
  \vspace{-3ex}
  \label{fig:RA_SA_evolution}
\end{figure}

Previous studies on video-based OSA detection have focused on  distinguishing subtle differences in chest-area motion between obstructive apnea and normal breathing~\cite{munoz2020home,abad2016automatic,akbarian2021noncontact}. However, these direct approaches suffer from low accuracy or substantial computational burden (20 hours of inference time). In contrast, we propose a novel problem conversion that aligns with clinical expertise, resulting in a more \textit{video-friendly task}. Specifically, we \textbf{indirectly detect OSA by identifying respiratory arousal (RA) events}, which are awakenings that can occur within 3 seconds of an apnea/hypopnea event. Our intuition is that RA events exhibit strong linear correlations with apnea/hypopnea events, and involve more noticeable and distinct motions.

After converting the problem from OSA to RA detection, our focus shifts to assessing the feasibility of differentiating RA events from  various other sleep events, including different types of arousals. To investigate the feasibility, we conduct an empirical analysis on the sleep video dataset. In this study, we calculate the pixel-wise differences between two frames, generating a \textit{difference image} that consists of \textit{difference pixels} to effectively focus on human motions and filter out noise. We define \textit{large difference pixels (LDPs)} as difference pixels with values exceeding a threshold.

Figure \ref{fig:Bar_all} illustrates the average number of LDPs per difference image for various sleep events, revealing distinct movement patterns in arousal events (RA, spontaneous arousal (SA), and periodic limb movement while awake (PLMA)) and wake stages compared to other sleep events (Apnea, Hypopnea, Desaturation). To explore the feasibility of distinguishing RA from other arousals and wake stages, we observe movements \textit{preceding} the arousals, with Figure~\ref{fig:PLM_RERA} showing clear differentiation between Wake and RA, as well as distinctive characteristics for PLMA. However, distinguishing SA from RA remains challenging according to similar shapes in their violin plots. Further analysis in Figure~\ref{fig:RA_SA_evolution} reveals a slightly but consistently lower average LDP count in pre-RA footage compared to pre-SA, indicating potential feasibility for differentiation. Nonetheless, the subtle difference between RA and SA poses challenges for simple rule-based algorithms, as confirmed by \textit{significant overlap} shown in shaded regions in Figure~\ref{fig:RA_SA_evolution}.

\vspace{1ex}\noindent
\textbf{Summary.} 
Overall, the empirical studies demonstrate that our strategy for \proposal, which indirectly detects OSA through RA, can be effective by easily filtering out most sleep events. At the same time, however, \proposal should involve a sophisticated mechanism that thoroughly analyzes \textit{long-term characteristics} of subtle sleep motions, especially to differentiate between RA and SA. 

\section{Method}
\label{sec:method}

\begin{figure*}[t]
  \centering
    \includegraphics[width=.9\textwidth]{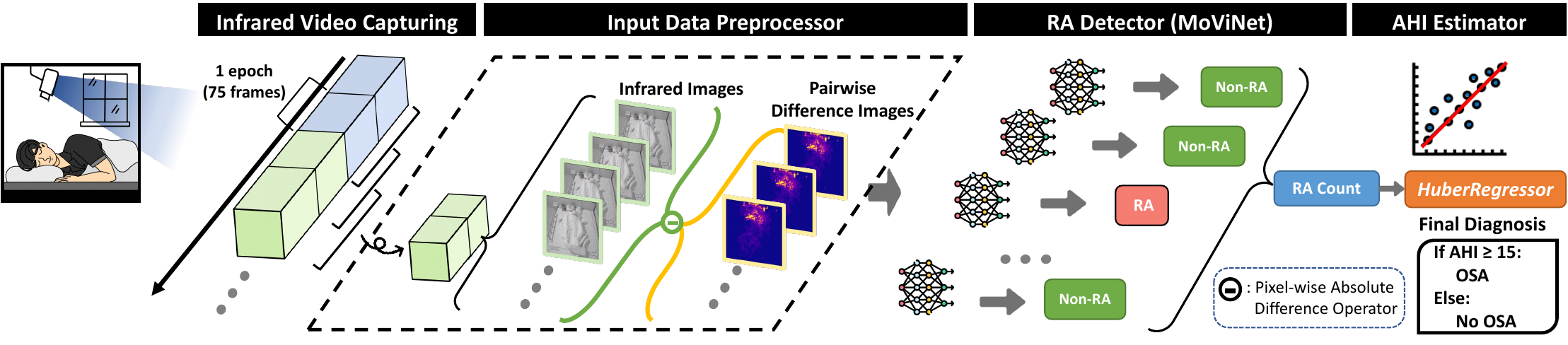}
    \vspace{-2ex}
  \caption{\proposal Architecture Overview }
  \vspace{-2.5ex}
  \label{fig:sys-architecture}
\end{figure*}

\proposal (Figure~\ref{fig:sys-architecture}) estimates AHI by counting the occurrence of RA during the total sleep time. It comprises Input Data Preprocessor, RA Detector, and AHI Estimator. 
%


\vspace{.5ex}\noindent
\textbf{Input Data Preprocessor.} 
The Input Data Preprocessor refines the original infrared video footage into an optimized input clip tailored to lightweight RA detection.
By using an appropriate clip size and sliding window step for the input data, we can accurately capture all RA events occurring throughout the entire sleep duration. As emphasized in \S\ref{key_insight}, one of the key distinguishing features of RA from other arousals, particularly SA, is the occurrence of apnea and hypopnea events preceding RA. Considering these observations and sleep medicine knowledge, we set the input data to 60 second-clips with a step size of 30 seconds. This ensures that the RA Detector effectively captures not only the movement patterns of RA but also the patterns of events preceding RA, allowing for successful differentiation. 
Furthermore, in order for the RA Detector to accurately capture subtle motions within the clips and operate robustly in the presence of various noise and environmental changes, appropriate preprocessing is required. By applying the frame difference method to still-like sleep videos, our model can focus solely on the motion, unaffected by variations in the background and noise present in different hospital settings.

\vspace{.5ex}\noindent
\textbf{Respiratory Arousal (RA) Detector.} 
The RA Detector is a DNN that takes a 1-minute clip of frame differences as input and performs binary classification to determine whether it belongs to the RA or not. We adopt MoViNet~\cite{kondratyuk2021movinets}, a recently developed efficient video recognition model, as the RA Detector. Specifically, we choose the MoViNet-A0 architecture with (2+1)D convolutions due to the advantage of low computational requirements and memory usage, making it highly suitable for on-device deployment. 

To effectively train the RA Detector, we curate the train dataset. Due to the characteristics of sleep videos where frames show minimal changes over short durations, extracting multiple clips from consecutive time zones may lead to redundant and uninformative clips. To address this, we  select a single 1-minute clip per 10-minute video. Moreover, when selecting RA clips, we ensure that the model observes at least 15 seconds of pre-arousal behavior, a crucial period for distinguishing RA from SA.

\vspace{.5ex}\noindent
\textbf{AHI Estimator.}
The AHI represents the number of apnea and hypopnea events per hour of actual sleep time, excluding periods of wakefulness before falling asleep or during sleep interruptions. Given the availability of total time spent in bed (TIB) rather than actual sleep time, we calculate the RA ratio by dividing RA events by TIB to estimate the AHI. Utilizing the linear relationship between RA counts and AHI, as discussed in \S\ref{key_insight}, we opt for a robust linear regression model, the Huber Regressor, to estimate the AHI, as it can effectively handle outliers~\cite{huber2011robust, maronna2019robust}. For OSA prediction, we determine the presence of OSA using an AHI threshold of AHI $\geq$ 15, which is widely adopted in research and clinical practice~\cite{senaratna2017prevalence, kim2019prediction, liu2022classification, kapur2017clinical} due to its significance in identifying individuals at increased risk of adverse health outcomes if left untreated~\cite{goyal2017obstructive, young2008sleep}.

\vspace{-1ex}
\section{Evaluation}
\vspace{-0.5ex}

\begin{table*}[t]
\renewcommand{\arraystretch}{0.95}
\centering
\caption{Performance of AHI Estimation and OSA prediction}
\label{tab:evalAHI}
\vspace{-2ex}
\resizebox{\linewidth}{!}{ 
{\tiny
\begin{tabular}{c|c|cc|cccc}
\hline
\toprule
\multirow{2}{*}{Dataset} & \multirow{2}{*}{Estimator fitting } &
 \multicolumn{2}{c|}{AHI Estimation} &  \multicolumn{4}{c}{OSA Prediction} \\ 
& dataset &
\begin{tabular}[c]{@{}c@{}} Spearman correlation \\ coefficient ($\rho$)\end{tabular} &
P-value & Accuracy (\%) & Precision & Recall & F1 Score \\
\hline
\midrule
A valid (50 patients) & 
A valid & 0.827 & 1.37e-13 & 84.0 
& 0.886 & 0.886 & 0.886 \\
A test (50 patients) & 
A valid & 0.744 & 5.89e-10 & 82.0 
& 0.842 & 0.914 & 0.876 \\
B test (115 patients) & 
A valid & 0.756 & 8.60e-23 & 83.4 
& 0.867 & 0.918 & 0.891 \\
C test (80 patients) & 
A valid & 0.834 & 8.16e-22 & 65.0 
& 1.000 & 0.594 & 0.745 \\
C test (80 patients) & 
A \& C valid & 0.834 & 8.16e-22 & 83.7 
& 0.924 & 0.884 & 0.903 \\
\bottomrule
\hline
\end{tabular}
}
} 
\vspace{-2ex}
\end{table*} 

\begin{table*}[t]
\renewcommand{\arraystretch}{1}
  \centering  
  \caption{Performance on Jetson Nano}
  \vspace{-2.5ex}
  \label{tab:on_device_evaluation}
\resizebox{\linewidth}{!}{ 
{\tiny
  \begin{tabular}{ccccccc}
  \hline
    \toprule
    \begin{tabular}[c]{@{}c@{}} Model size\\ (FP16)\end{tabular} &
    \begin{tabular}[c]{@{}c@{}} Model load\\ Frame capture ready\end{tabular} & 
    \begin{tabular}[c]{@{}c@{}} Frame processing\\ (1 min. clip)\end{tabular} & 
    Inference & Total Operation &
    \begin{tabular}[c]{@{}c@{}} Peak Memory \\ (RSS)\end{tabular} &
    \begin{tabular}[c]{@{}c@{}} Peak Memory \\ (Runtime)\end{tabular}  \\
    \hline
    \midrule
    5.1 (MB) & 
    1.035$\pm$0.007 (s) & 0.224$\pm$0.042 (s) & 3.040$\pm$0.046 (s) & 3.264$\pm$0.088 (s) &
    839$\pm$15.5 (MB) &2.67$\pm$0.016 (GB) \\
  \bottomrule
  \hline
  \end{tabular}
}
} 
\vspace{-3ex}
\end{table*}

\noindent
\textbf{RA Detector.} 
The RA Detector is trained on 449 patients’ clips and evaluated on 50 patients' clips from Hospital A. Within the training data, clips from 50 patients were set aside
for validation purposes, enabling 9-fold cross-validation.
We employ the Area under the curve (AUC) of the Receiver Operating Characteristic (ROC) curve as the evaluation metric, widely used in medical diagnostic tests~\cite{hajian2013receiver}. The RA Detector achieves a mean AUC of 0.79 with 9-fold cross-validation, which is considered nearly \textit{excellent}~\cite{hosmer2013applied, mandrekar2010receiver}, indicating its high discriminatory power in distinguishing between RA and non-RA clips.

\vspace{1ex}\noindent
\textbf{AHI Estimator.}
To assess the strength of the relationship between the estimated AHI and the ground-truth AHI measured by the gold standard PSG, we conduct Spearman correlation analysis, which is a rank-order correlation. 
The AHI Estimator is evaluated on the test set from three different institutions, showing robust performance with high Spearman correlation coefficient $\rho$ values between the estimated AHI and PSG AHI (Table \ref{tab:evalAHI}).

\vspace{1ex}\noindent
\textbf{OSA prediction.}
We evaluate the OSA prediction performance using metrics such as accuracy, precision, recall, and the F1 score. The system demonstrates robust performance for Hospitals A and B, achieving accuracies of 82.0\% and 83.4\% and F1 scores of 0.876 and 0.891, respectively (Table \ref{tab:evalAHI}). However, the performance for Hospital C was relatively lower, which can be attributed to environmental variations, such as a thicker blanket and darker video footage in Hospital C. To address this, the AHI Estimator was refined by re-fitting it using the A valid set and incorporating 35 additional cases from the C valid set. This tailored approach improved performance for Hospital C, resulting in an accuracy of 83.7\% and F1 score of 0.903. The approach effectively addresses the challenges posed by environmental variations, enhancing the overall robustness of the system.
In addition, although the current dataset provides only single-night data for each patient, aggregating multi-night results in real use cases can improve detection performance further.

\vspace{1ex}\noindent
\textbf{On-device Performance.} 
Table \ref{tab:on_device_evaluation} presents the performance of \proposal on the resource-constrained Jetson Nano (quad-core ARM A57 CPU and 4 GB 64-bit LPDDR4 memory). The initialization step takes $\sim$1.045 seconds, while the Input Data Preprocessor requires $\sim$0.224 seconds to process frames from a 1-minute clip. The RA Detector takes 3.04 seconds for inference, resulting in \textbf{$\sim$3.264 seconds on CPU} for determining RA from a 1-minute clip.  In terms of peak
memory usage, this process takes up to about 839 MB in Resident
Set Size (RSS) and about 2,730 MB in runtime.
The system achieves real-time results, allowing a  resource-constrained device to not store many video frames. Alternatively, individuals can input separately recorded sleep videos for retrospective analysis, taking $\sim$50 minutes for an 8-hour video. The inference speed is \textbf{$\sim$19$\times$ faster} than the previous method~\cite{akbarian2021noncontact}, irrespective of resource constraints.

\begin{figure}[t]
  \centering
  \vspace{-0.5ex}
    \includegraphics[width=\linewidth]{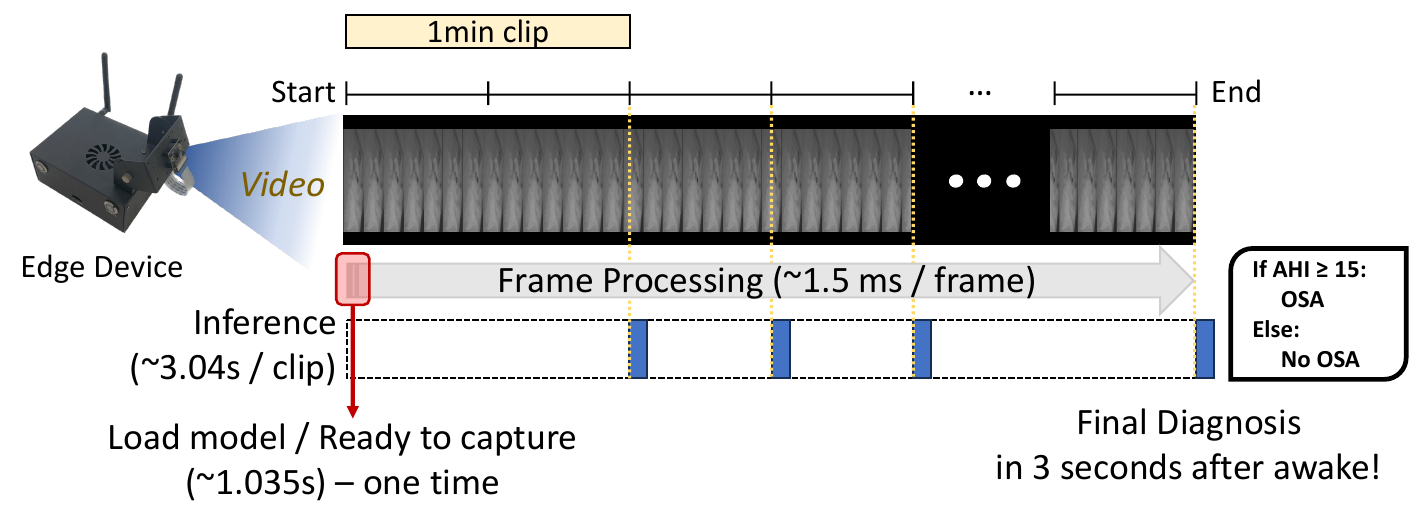}
    \vspace{-5ex}
  \caption{On-device Operation Overview}
  \vspace{-3ex}
  \label{fig:jetson}
\end{figure}

\vspace{-1ex}
\section{Conclusion}
\vspace{-0.5ex}

This work is the first systematic exploration on non-contact, on-device OSA diagnosis using infrared video. By leveraging clinical expertise and deeply analyzing the largest sleep video dataset to date, we have established a novel strategy: indirectly detecting OSA through the identification of RA events.
The combination of this indirect strategy, tailored design of input data and training data, and lightweight DNN has enabled real-time on-device inference for detecting  subtle sleep motions.  
With the identification of infrared video as an effective data type for sleep monitoring, we believe that this work can serve as a stepping stone to advance the field of sleep disorder diagnostics and enhance the accessibility of sleep medicine.

\vspace{1ex}\noindent
\textbf{Acknowledgement}
This work was partially supported by the National Research Foundation of Korea (NRF) grant funded by the Korea government (MSIT)(No. RS-2023-00212780), Creative-Pioneering Researchers Program through Seoul National University, the SNUH Research Fund (grant no 0320222200), and was a part of the 'AI Dataset Project' (aihub.or.kr), funded by the Ministry of Science \& ICT and the National Information Society Agency, Republic of Korea.

{\small
\bibliographystyle{ieee_fullname}
\bibliography{main}
}

\end{document}